\documentclass[letterpaper, 10 pt, conference]{ieeeconf}  % 
 \pdfoutput=1

\IEEEoverridecommandlockouts % This command is only needed if 
\usepackage{amsmath,amsfonts}
\usepackage{algorithmic}
\usepackage{array}
\usepackage[caption=false,font=normalsize,labelfont=sf,textfont=sf]{subfig}
\usepackage{textcomp}
\usepackage{stfloats}
\usepackage{url}
\usepackage{verbatim}
\usepackage{graphicx}
\usepackage{layouts}
\usepackage[acronym]{glossaries}
\usepackage{float}
\usepackage{tabularx}

\title{\LARGE \bf
Semi-Automatic Infrared Calibration for Augmented Reality Systems in Surgery*
}

\author{Hisham Iqbal$^{1}$ and Ferdinando Rodriguez y Baena$^{1}$% <-this % stops a space
\thanks{$^{*}$This work was supported by the EPSRC in the form of a Knowledge Transfer Secondment (project reference: EP/R511547/1)}% <-this % stops a space
\thanks{$^{1}$H. Iqbal and F. Rodriguez y Baena were with the Mechatronics in Medicine Laboratory, Dept. of Mechanical Engineering, Imperial College London. {Email: \tt\small hisham.iqbal13\,[at]\,ic.ac.uk}
}%
}

 \makeatletter
\let\NAT@parse\undefined
\makeatother

\usepackage{hyperref}
\hypersetup{
colorlinks,
linkcolor=blue, 
urlcolor=blue,
filecolor=blue,
citecolor=blue
}

\begin{document}
 \thispagestyle{empty}
 
\begin{figure*}
{\LARGE \bf IEEE Copyright Notice}
\vspace{3cm}

\copyright\ 2022 IEEE. Personal use of this material is permitted. Permission from IEEE must be
obtained for all other uses, in any current or future media, including reprinting/republishing
this material for advertising or promotional purposes, creating new collective works, for resale
or redistribution to servers or lists, or reuse of any copyrighted component of this work in other
works.
\newline
\newline
Accepted for publication in: \textbf{2022 IEEE/RSJ International Conference on Intelligent Robots and Systems (IROS)}
\newline
\textbf{DOI:} \href{https://doi.org/10.1109/IROS47612.2022.9982215}{10.1109/IROS47612.2022.9982215}
\vspace{12cm}

\noindent\rule[7pt]{\linewidth}{0.4pt}
For associated code visit: \href{https://github.com/HL2-DINO}{https://github.com/HL2-DINO}
\end{figure*}
\newpage
\maketitle
 \thispagestyle{empty}
\pagestyle{empty}
%%%%%%%%%%%%%%%%%%%%%%%%%%%%%%%%%%%%%%%%%%%%%%%%%%%%%%%%%%%%%%%%%%%%%%%%%%%%%%%%
\begin{abstract}
Augmented reality (AR) has the potential to improve the immersion and efficiency of computer-assisted orthopaedic surgery (CAOS) by allowing surgeons to maintain focus on the operating site rather than external displays in the operating theatre. 
Successful deployment of AR to CAOS requires a calibration that can accurately calculate the spatial relationship between real and holographic objects. Several studies attempt this calibration through manual alignment or with additional fiducial markers in the surgical scene.
We propose a calibration system that offers a direct method for the calibration of AR head-mounted displays (HMDs) with CAOS systems, by using infrared-reflective marker-arrays widely used in CAOS. In our fast, user-agnostic setup, a HoloLens 2 detected the pose of marker arrays using infrared response and time-of-flight depth obtained through sensors onboard the HMD. Registration with a commercially available CAOS system was achieved when an IR marker-array was visible to both devices.
Study tests found relative-tracking mean errors of 2.03 mm and 1.12$^{\circ}$ when calculating the relative pose between two static marker-arrays at short ranges. When using the calibration result to provide in-situ holographic guidance for a simulated wire-insertion task, a pre-clinical test reported mean errors of 2.07 mm and 1.54$^{\circ}$ when compared to a pre-planned trajectory. 
\end{abstract}


%%%%%%%%%%%%%%%%%%%%%%%%%%%%%%%%%% INTRO
\newacronym{CAOS}{CAOS}{computer-assisted orthopaedic surgery}
\newacronym{AR}{AR}{augmented reality}
\newacronym{HMD}{HMD}{head-mounted display}
\newacronym{PFA}{PFA}{patellofemoral arthroplasty}
\newcommand{\navio}{NAVIO\textregistered}

\section{Introduction}
\label{section:intro}
The development and increased adoption of \gls{CAOS} has provided a pathway for improving precision in follow-up outcomes in joint replacement procedures \cite{Shatrov2020ComputerOutcomes} (e.g. limb and prosthetic alignment). The foundation of CAOS systems is their navigation hardware, which enables tracking of relative motion between a patient and surgical tools. The industry standard for navigation in CAOS is the use of stereo infrared (IR) cameras, which calculate the 6 DOF pose of IR-reflective marker-arrays using image processing and stereo triangulation \cite{Mezger2013NavigationSurgery}. 

Despite the clinical maturity of CAOS, there are still potential improvements to be made with the presentation of medical data and handling of user interactions. Medical data that is generated or used intraoperatively in CAOS is presented to the surgeon in the 2D frame of a touchscreen monitor, despite its 3D nature (patient scans, implant models etc.). Cognitive challenges can arise from this, as the surgeon's viewpoint is not strictly aligned with the monitor's displayed orientation, requiring a mental transformation between two coordinate frames, consequently directing focus to a display and away from the operating site. Augmented reality (AR) – a technique of superposing computer-generated imagery with the real environment, presents a method to address this and help return the surgeon's focus back to the operating site.    
In recent years, researchers have explored using \glspl{HMD} such as the HoloLens 1 \& 2 (Microsoft Corporation, WA, USA) and Magic Leap One (Magic Leap Inc., FL, USA) to meet this challenge, by using AR to render holographic content directly on the operating site in meaningful locations with respect to patient anatomy. Potential use cases for AR in orthopaedic scenarios have been comprehensively reviewed by Jud et al. \cite{Jud2020ApplicabilityReviewb}. 
Some examples of studies investigating AR-guidance for assisting orthopaedic surgical tasks include (and are not limited to) guidewire positioning for shoulder arthroplasty \cite{Kriechling2021AugmentedStudy}, pedicle screw implantation \cite{Muller2020AugmentedImaging}, and guiding needle-placement for lumbar facet joint injections \cite{Agten2018AugmentedInjections}.

Successfully integrating an HMD in the loop of an existing CAOS robot's workflow requires a calibration between an HMD and any surgical tracking systems used by the robot (e.g. optical trackers). As modern HMDs are capable of inside-out tracking (localising the headset's position and orientation in the world), it is possible to calculate a registration between the real coordinate frame of a tracking device, and the virtual coordinate frame of an HMD. A review by Andrews et al {\cite{Andrews2021RegistrationDevices}} provides detail on several techniques reported in literature that address the registration challenge for AR setups targeting clinical applications.

\begin{figure*}
\centering
\vspace*{4px}
    \includegraphics[width=17cm]{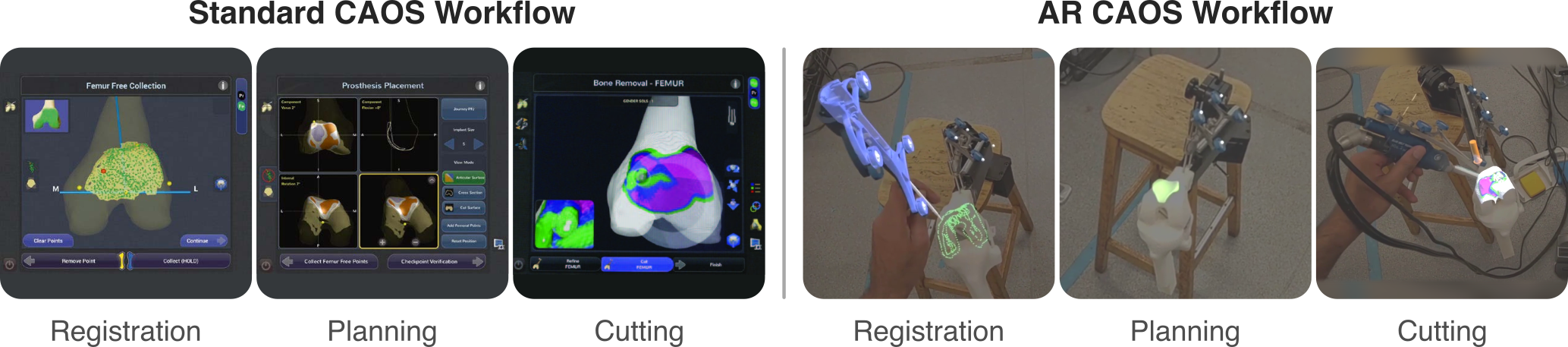}
    %\glsreset{PFA}
    \caption{Comparison of workflows designed for computer and robot-assisted patellofemoral arthroplasty (PFA). Left: screenshots of the standard, monitor-based workflow run by the {\navio} robot -- a commercial CAOS system. Right: Screenshots of our novel AR-centric workflow which reproduces the {\navio} robot's entire surgical workflow for PFA in a 3D AR environment through an HMD {\cite{Iqbal2022AugmentedSurgery}} -- an application of the AR-registration method described in this paper.} 
    \label{fig:ar_caos}
    \vspace*{-8px}
\end{figure*}
Multiple studies in this field employ manual techniques, which use variants of the single-point active alignment method (SPAAM) \cite{Tuceryan2002Single-PointReality}, based on instructing users to align a real, physical object with virtual points rendered on an HMD. Provided the `real' object can be externally tracked and the positions of the virtual points are known, it is possible to determine a correspondence between the coordinate frame of a tracking device and the virtual coordinate frame of the headset. Various methods and techniques reported in literature cite improved SPAAM setups for more user-specific calibration \cite{Hu2021Rotation-constrainedAlignment, Azimi2017AlignmentOT}, however, an underlying issue of manual-alignment techniques is a reliance on subjective user input and time-consuming calibration procedures. 
An alternative approach is to use front-facing visible-light cameras (VLCs) on HMDs for image-marker (e.g. ArUco or QR markers) tracking. Open-source projects integrating libraries such as OpenCV and Vuforia (PTC Inc., MA, USA) are becoming more commonplace, and provide resources for creating HMD compatible applications for image-marker tracking scenarios. Image-marker tracking helps remove the subjectivity of manual alignment, but when used to calibrate HMDs and surgical tracking systems, it becomes necessary to manufacture custom `hybrid' calibration tools. These tools are designed to be equipped with both image- and IR-reflective markers \cite{8769846}, making them visible to both VLCs and optical trackers. Following the manufacture of a hybrid calibration rig, an additional offline calibration is required to calculate the geometric transformation between the image- and IR-reflective markers, the estimation of which can introduce additional sources of error during subsequent calibration.   

In contrast, emerging research has explored using AR-HMDs for inside-out tracking of surgical instruments equipped with retroreflective IR markers, which are commonly utilised in computer and robot-assisted surgery. In recent years, HMD manufacturers such as Microsoft have released open-source projects providing direct access to sensor streams {\cite{Ungureanu2020HoloLens2R}} for HMDs such as the HoloLens 1 \& 2, which can facilitate these goals. 
Gsaxner et al. {\cite{Gsaxner2021Inside-outReality}} explored detecting and tracking IR-reflective marker arrays by attaching an infrared emitting source to the HoloLens 2, and analysing paired grayscale stereo-images returned by the headset's front-facing visible light cameras. Kunz et al. {\cite{Kunz2020InfraredInterventions}} explored similar techniques and also used the HoloLens 1 to track IR-reflective markers by processing IR response and time-of-flight (ToF) depth returned by the HMD's front-facing depth sensor. However, to the best of the authors' knowledge, there is a lack of studies in literature that aim to co-register AR-HMDs and surgical tracking systems utilising similar techniques – solely with the use of IR-reflective marker arrays.

Thus, we sought to build on the existing research landscape for AR-registration with surgical tracking systems, by constructing a calibration setup exploiting the IR tracking sensors onboard a HoloLens 2 to simplify the task of registration, whilst eliminating drawbacks of manual or marker-based techniques such as subjective user-input, additional system modifications and/or tracking infrastructure. Our system aimed to facilitate easier integration of AR-HMDs with established CAOS systems, to simplify the deployment of AR in computer and robot-assisted surgical workflows, and to use clinically established optical trackers to guide in-situ positioning of holographic content.

Fuelled by early evidence that AR could provide benefits in {\gls{CAOS}} workflows {\cite{Iqbal2021AugmentedStudyb}}, we have been working on augmenting the workflow of a commercial CAOS platform, the {\navio} (Smith \& Nephew plc., UK), with a mixed-reality workflow co-registering optically tracked patient-anatomy with virtual content {\cite{Iqbal2022AugmentedSurgery}} (as seen in Figure {\ref{fig:ar_caos}}). In this paper, we detail the calibration system which enabled the integration of AR into the surgical workflow of this commercial CAOS system. Our setup made no additional modifications to the HMD, and employed a user-agnostic registration algorithm, where a user wearing the HMD looked at a target array of IR markers visible to both the HoloLens 2 and {\navio} robot, and co-registered the holographic coordinate frame of the HMD with the coordinate frame of the surgical robot's optical tracker in a matter of seconds. No additional tracking infrastructure was introduced when running the workflow for a \navio\ assisted patellofemoral arthroplasty (PFA), and the presented system used a hybrid approach to track surgical tools and patient anatomy with either the HoloLens 2 or the \navio's optical tracker. Whilst the presented setup was integrated with the {\navio} robot, its design can be generalised and applied to calibrate a HoloLens 2 with any commercially available stereo infrared camera. 

An overview of the system design is included within Section \ref{section:matMethods}, providing detail on how sensor data obtained from the HoloLens 2 can be used to locate infrared markers, and in turn, register the position and orientation of marker-arrays tracked by a surgical robot. The results of a pre-clinical quantitative evaluation of the system are included in Section \ref{section:results}, where the achieved registration is evaluated with two tests: (i) relative tracking errors when using the HoloLens 2 to track multiple IR marker arrays (ii) surgical task error when carrying out an AR-assisted phantom-drilling task. The paper concludes with a discussion of the results and study limitations.

%%%%%%%%%%%%%%%%%%% MATMETHODS

\begin{figure*}
    \vspace*{8px}
    \includegraphics[width=\linewidth]{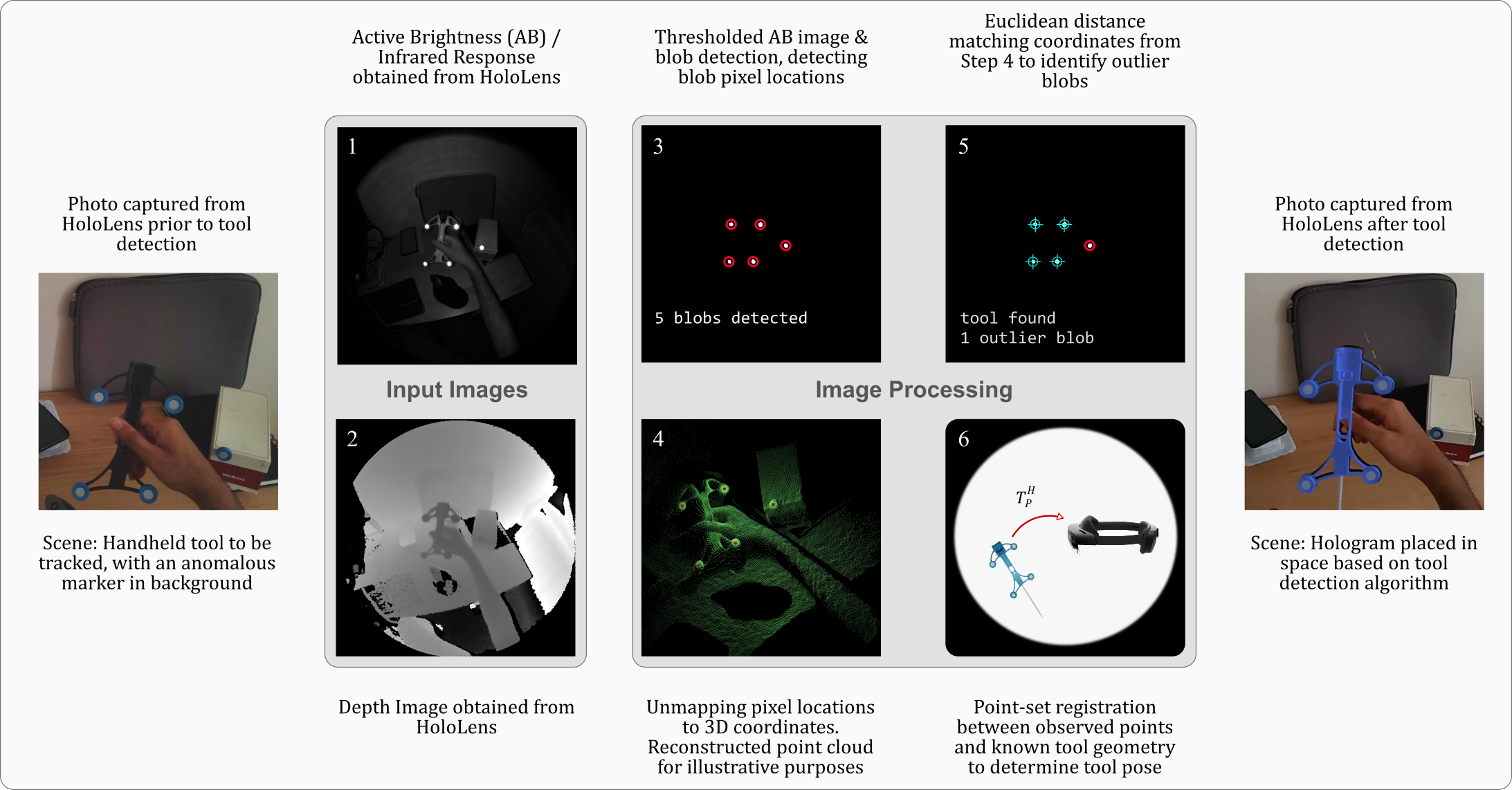}
    \caption{ Pipeline for image processing of sensor images for tool detection and tool pose estimation. \textbf{Left}: Scene prior to tool detection. \textbf{(1)} Input active-brightness (AB) image. \textbf{(2)} Input depth image. \textbf{(3)} Processed AB image following thresholding \& blob detection. \textbf{(4)} Reconstructed point cloud from depth image in Figure \ref{fig:img_proc}.2, points coloured by IR response. \textbf{(5)} Outlier detection of blobs after unmapping pixel-coordinates to 3D-coordinates and comparing with known tool geometries. \textbf{(6)} Registration process to calculate pose of tool with respect to virtual world frame. \textbf{Right}: Scene after tool detection and AR augmentation.}
    \label{fig:img_proc}
    \vspace*{-8px}
\end{figure*}

\section{Materials and Methods}
\label{section:matMethods}

\subsection{Sensor Datastreams}
\label{subsection:sensor-datastreams}
By utilising the Research Mode API  \cite{Ungureanu2020HoloLens2R} with the HoloLens 2, it is possible to obtain a data-stream to the array of sensors mounted to the headset. The sensor array on the headset includes 4 visible-light grayscale tracking cameras, and two time-of-flight depth sensors; a near-depth articulated hand-tracking (AHAT) sensor and a long-range depth sensor used for spatial mapping. 

Of the two depth sensors, the AHAT sensor was determined as a more suitable candidate for this project due to its higher frame rate (45 FPS vs 5-10 FPS \cite{Ungureanu2020HoloLens2R}). Each frame, the AHAT sensor returned two image buffers which were used as inputs to this system:

\begin{enumerate}
    \item \textbf{Active Brightness (AB) Image}: A grayscale image showing infrared response of the scene
    \item \textbf{Depth Image}: A grayscale time-of-flight depth image indicating depth of pixels seen in the paired AB image
\end{enumerate}

Figure \ref{fig:img_proc} contains examples of both the AB and depth images that are provided by the headset. The AB and depth images were two modes of the headset's near-depth IR stream, and were computed from the same modulated IR signal for depth computation {\cite{Ungureanu2020HoloLens2R}}. Both images had a resolution of $512 \times 512$ pixels, containing unsigned 16-bit integers. A constraint placed by the hardware and API was that information obtained from the depth image returned was only valid in a 1-metre range from the headset.  

Based on the images obtained from the AHAT sensor -- depth and infrared response, the tracking of tools (equipped with IR-reflective markers) and calibration of the HoloLens 2 with the \navio's optical tracker was achieved by carrying out the following tasks: 

\begin{itemize}
    \item \textbf{AB Image Processing:} The AB image is processed to find pixel locations of IR reflective markers
    \item \textbf{Depth Image Processing:} The depth values at these pixel locations in the depth image are grabbed and converted into camera-centric 3D positions
    \item \textbf{Calculating Tool Pose:} The computed 3D positions are compared against known tool geometries to remove outliers, and then used to calculate the pose of any detected tools with respect to a fixed virtual coordinate frame (defined as the startup location of the HoloLens 2)
    \item \textbf{HoloLens 2 \& Optical Tracker Calibration:} The fixed virtual coordinate frame of the HoloLens 2 is calibrated with the optical tracker's coordinate frame, by using tools that are visible to both devices
\end{itemize}

\subsubsection{\textbf{AB Image Processing}}
\label{subsection:ABImgProc}
As seen in Figure \ref{fig:img_proc}, the presence of infrared markers produces saturated bright spots in the AB image ($AbImg$ in Equation \ref{eqn:ab-img-thresh}) at a variety of distances. The raw AB frames are passed through a binary thresholding mask to isolate the brightest pixels in a frame. 
\begin{equation}
\label{eqn:ab-img-thresh}
        AbImg(u,v)=
    \begin{cases}
      AbImg(u,v), & \text{if}\ AbImg(u,v) > thresh \\
      0 & \text{otherwise}
    \end{cases}
\end{equation}
Thresholding can provide false negatives for regions of interest, as objects that are in very close proximity to the sensor can reflect a high proportion of the IR light back. Consequently, the second stage of the process is running a blob filter that isolates contours based on circularity, area and convexity. This two-stage processing now means a raw AB image can be processed to search for round, bright spots likely to be associated with the presence of infrared markers.
The output of this first image processing step is a list of $n$ candidate keypoint `blob' pixels for marker centres ($\{(u_i, v_i),...,(u_n, v_n)\}_{i = 1..n}$), which are passed onto the next step of the workflow.

\subsubsection{\textbf{Depth Image Processing}}
\label{subsection:DepthImgProc}

The candidate `blob' pixels isolated from the AB image \{$(u_i, v_i),...$\}, are used to query the corresponding ToF depth image to obtain an associated depth value ($d_i$) relative to the sensor coordinate frame $C$, for each pixel location.

In order to convert pixel locations to a 3D coordinate $^c{\mathbf{p}}$, the Research Mode API is queried to unmap a pixel coordinate ($u_i, v_i$) to a normalised 3D position ($^c\mathbf{\hat{p}}$), which is then scaled by its associated scalar depth value 
\begin{gather}
^c\mathbf{{p}}_i = d_i \times ^c\mathbf{\hat{p}}_i 
\end{gather}

The resulting 3D coordinate is relative to the depth sensor's coordinate frame $C$, $^c\mathbf{{p}}_i = [x_i, y_i, z_i]'$. As the HoloLens 2 is capable of inside-out tracking, the software then queries the sensor's API each frame for an extrinsic matrix $T_{cam2World}$. This extrinsic matrix can then be used to map coordinates from the sensor coordinate frame $C$ to a virtual coordinate frame $W$.

Repeating this for each candidate pixel produces a list of $n$ unordered 3D points \{$^w{\mathbf{p}_i},...,^w{\mathbf{p}_n}$\} corresponding to IR-reflective marker centres in the virtual world-frame, which is passed to the next stage for calculating tool pose.

\begin{figure}
    %\centering
    \vspace*{8px}
    \includegraphics[width=\linewidth]{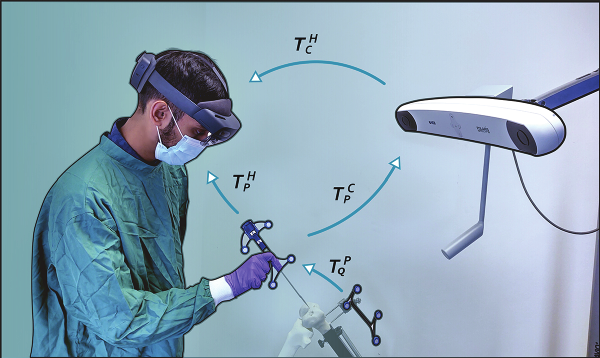}
    \caption{Various transformation matrices generated when using system: tool pose in fixed virtual world frame, $T^H_P$, tool pose with respect to surgical robot's optical tracker, $T^C_P$, and calibration matrix which maps between optical tracker and virtual coordinates, $T^H_C$. $T^P_Q$, matrix describing relative pose between two IR marker-arrays.}
    \label{fig:transform_diagram}
        \vspace*{-6px}
\end{figure}

\subsubsection{\textbf{Calculating Tool Pose}}
\label{subsection:CalcToolPose}
By comparing the unordered point-set data obtained from the previous step to known geometries of the tools tracked by the surgical robot, the problem can be framed as a point-correspondence and registration task. A fast Euclidean distance-matching algorithm was used for this paper to match and order the two point sets: observed IR marker coordinates, and known tool geometry coordinates. If no valid correspondence was found for a particular tool from the observed point-set, it was assumed the tool was not visible to the HoloLens 2. Otherwise, the now ordered point-set and known tool geometries were fed into a widely used least-squares point-set registration algorithm  \cite{Arun1987Least-squaresAnal}. Thus, if a tool was visible, it was possible to calculate its pose with respect to the virtual world frame of the HoloLens 2, $T_{toolToWorld}$ (e.g. ${T^H_P}$ in Figure \ref{fig:transform_diagram}).   
The tool's calculated pose was fed into a Kalman filter (which used experimentally determined parameters) to reduce the influence of noise in further processing. 

\begin{figure}

    \centering
    \includegraphics[width = 0.95\linewidth]{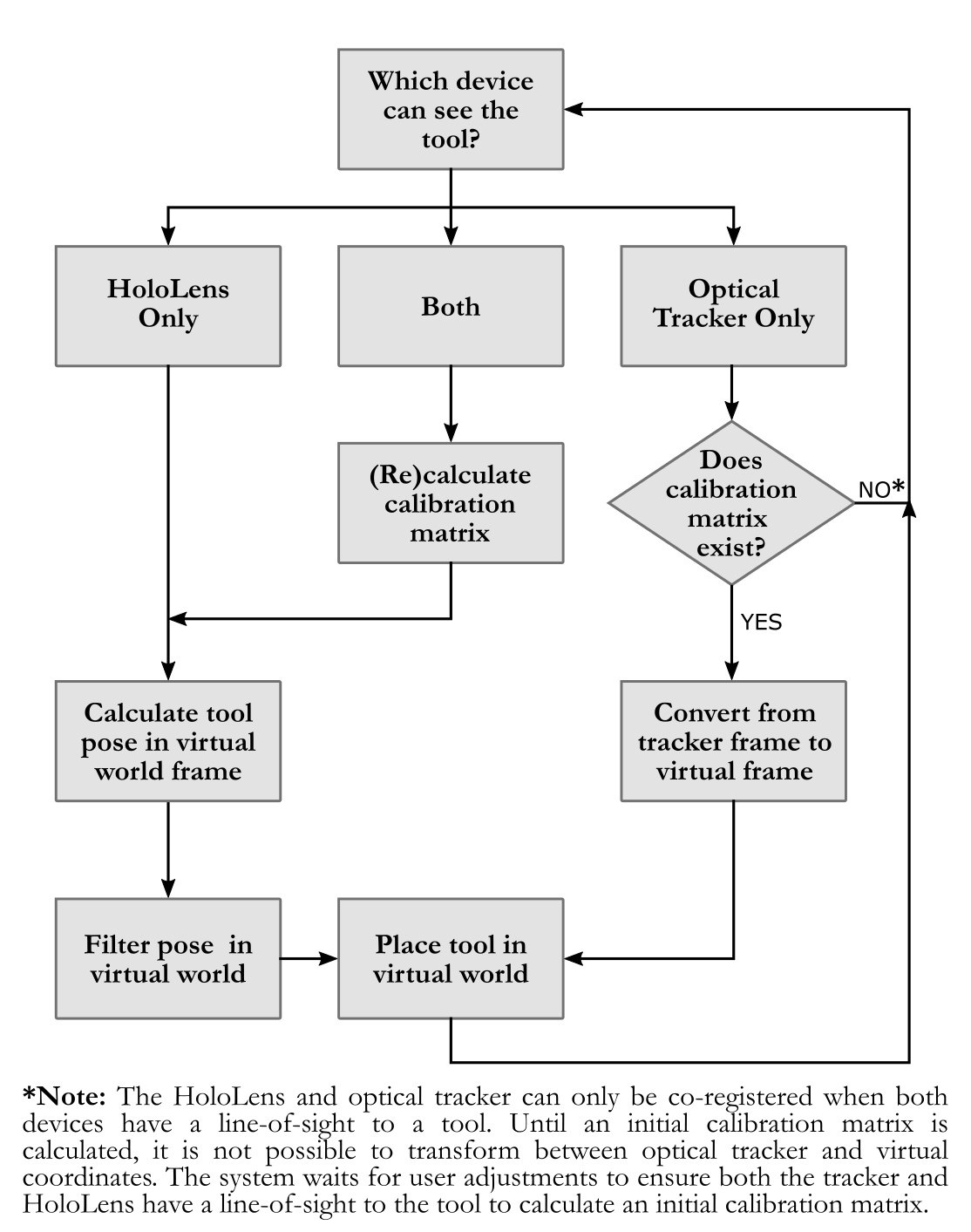}
    \caption{Flowchart illustrating how the system can track tools equipped with IR-reflective markers.}
    \label{fig:calibFlowchart}
      \vspace*{-8px}
\end{figure}

\subsubsection{\textbf{HoloLens 2 \& Optical Tracker Calibration}}
\label{subsection:SystemCalibration}
The final stage of the algorithm is to calibrate the two coordinate frames of the headset and surgical robot's optical tracker, which can be achieved if a tool or marker array is visible to both devices. During this phase, users were asked to avoid sudden, large head movements to avoid interfering with the HMD's self-locating processes. As illustrated in Figure \ref{fig:transform_diagram}, a tool's pose with respect to the optical tracker $(T^C_P)$ and its pose with respect to the HoloLens 2's fixed world frame $(T^H_P)$ can be used to calculate a rigid calibration matrix $(T^H_C)$ which maps between the optical tracker frame $C$ and holographic coordinate frame $H$ as follows:
\begin{equation}
\label{eqn:calibration-equation}
T^H_C = T^H_P \cdot (T^C_P)^{-1}
\end{equation}
With the calibration established, it was now possible to render and position virtual content based on optical tracker data. The system logic of how this calibration result was updated and used is illustrated in Figure \ref{fig:calibFlowchart}. In summary, provided a surgical tool or anatomy was visible to either the HoloLens 2 or optical tracker, a Unity (Unity Technologies, CA, USA) app running on the HoloLens 2 was able to render holographic augmentation in its virtual frame based on the tracked object's pose.

\subsection{System Setup}
\label{subsection:system-setup}
\begin{figure}
    \vspace*{3px}
    \centering
    \includegraphics[width = 7cm]{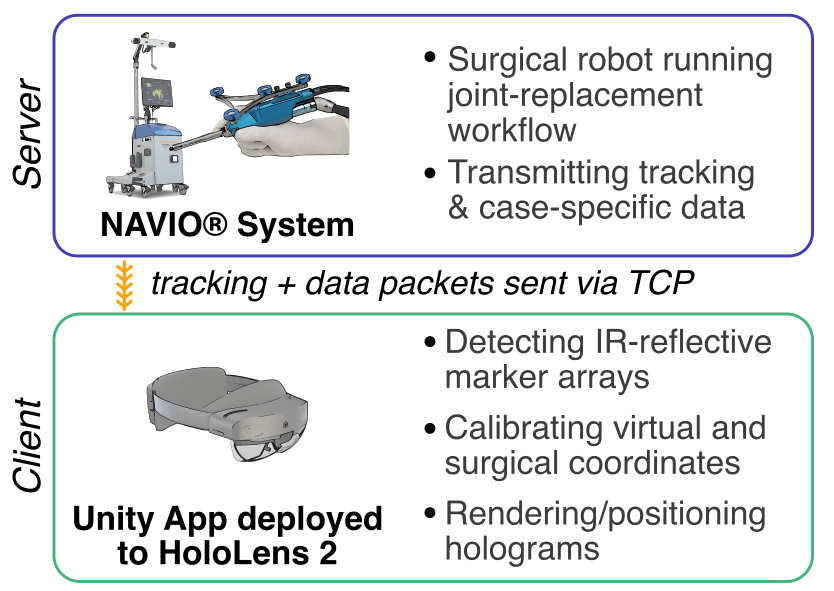}
    \caption{Diagram of system-flow between a HoloLens 2 headset and a custom build of the \navio\ robot's software}
    \label{fig:system-setup}
      \vspace*{-6px}
\end{figure}

The overall system structure (illustrated in Figure \ref{fig:system-setup}) consisted of two devices: a client HoloLens 2 device, communicating with a server software running on the surgical robot. 
On the client HoloLens 2 side, a Unity application was running which used a custom C++ DLL to interact with the Research Mode API of the headset. The DLL processed images on the HoloLens 2 as seen in Figure \ref{fig:img_proc}, and provided the pose of any tracked marker-arrays visible to the HMD. 

On the server-side, a bespoke build of the \navio\ software was developed in the context of a joint research collaboration with Smith \& Nephew (Innovate UK, 103950). The customised software provided continuous access to tracking packets supplied by the system's optical tracker, thus supplying pose for tracked tools and anatomy in the surgical tracker frame. The optical tracking packets were sent wirelessly over TCP to the HoloLens 2, where this information was parsed and used to carry out the calibration process described in Equation \ref{eqn:calibration-equation}.

\subsection{Experimental Setup}
\label{subsection:experiments}
A test framework was created to evaluate the proposed calibration system with the following tests: 
\subsubsection{Relative-Tracking Error}
\label{subsection:relative-tracking-experiment-description}
The first test aimed to evaluate the tracking error generated during the tool-pose estimation process (see Figure \ref{fig:img_proc}) on the HoloLens side. This was achieved by placing two stationary IR marker arrays in the field such that they were visible to both the HoloLens 2 and the surgical robot's optical tracker. If two tools (e.g. tool $P$ and tool $Q$), were visible to the HoloLens 2, $H$, the process outlined in Figure \ref{fig:img_proc} calculated their pose in the virtual world frame -- $T_P^H, T_Q^H$. This information was used to calculate an estimate for the relative pose between the two marker arrays (see $T^P_Q$ in Figure \ref{fig:transform_diagram}) as follows:
\begin{equation}
\label{eqn:measured-relativeT}
    {T_Q^P}_{Measured} = (T_P^H)^{-1} \cdot T_Q^H
\end{equation}
The ground truth to compare this result against was based on data provided by the surgical robot's optical tracker, $C$. A relative pose matrix ${T_Q^P}_{Actual}$ (see Figure \ref{fig:transform_diagram}) was calculated based on the pose of the two marker arrays in the tracker frame --  $T_P^C, T_Q^C$.
\begin{equation}
\label{eqn:actual-relativeT}
    {T_Q^P}_{Actual} = (T_P^C)^{-1} \cdot T_Q^C
\end{equation}
To maintain visibility to the HoloLens 2, the static experimental setup used marker-arrays placed within a 1 m distance of the headset (due to the constraint discussed in Section \ref{subsection:sensor-datastreams}). By comparing the measured and ground truth relative pose matrices as calculated in Equations \ref{eqn:measured-relativeT} - \ref{eqn:actual-relativeT}, it was possible to calculate a 6 DOF error (translation and rotation). Paired readings of the relative pose between two IR-marker arrays as calculated by the HoloLens 2 and optical tracker were used to compute the mean absolute relative tracking error associated with the system described in this paper.
\subsubsection{Surgical Task Error}
\label{subsection:relative-surgtask-experiment-description}

A second test was set up with the aim of evaluating task error when using holographic guidance to assist with a simulated surgical wire insertion task into the distal femur. The objective of the task was to align a tracked medical probe to a specified target position and orientation with respect to a plastic femur, thus aligning the tool with a pre-planned drilling trajectory. A virtual equivalent of the planned tool-trajectory was rendered in the HoloLens 2's coordinate frame by converting surgical coordinates to virtual coordinates using the calibration matrix obtained in Equation \ref{eqn:calibration-equation}. Eight volunteers (see Table \ref{table:relativeError} for metadata) were recruited to carry out the simulated surgical task described above. The inclusion criteria required the ability for volunteers to provide their own consent to participate as well as being above the age of 18. This single-centre study received ethical approval from the Imperial College Research Governance and Integrity Team (SETREC ref: 21IC6690).  

\begin{figure}
    \centering
    \vspace*{4px}
    \includegraphics[width=\linewidth]{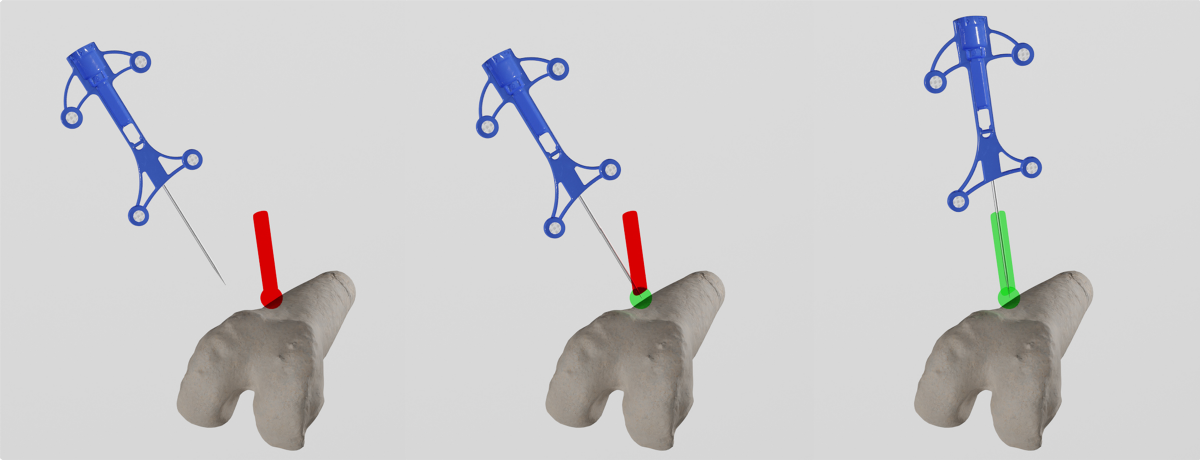}
    \caption{Illustration of holographic feedback shown to users during wire insertion task.  
    Colour of holographic cylinder and sphere modulate from red to green based on orientation and position errors respectively. Left: incorrect position \& orientation. Middle: correct position, but incorrectly oriented. Right: correctly positioned and oriented. }
    \label{fig:trackingScenarios}
      \vspace*{-6px}
\end{figure}

For each trial, volunteers followed the protocol outlined below:
\begin{itemize}
    \item Wear HoloLens 2 and follow standard device eye-calibration instructions
    \item Launch Unity app, gaze at a stationary marker-array visible to both headset and surgical robot to achieve system calibration
    \item Virtual drilling axis is placed relative to plastic femur anatomy based on system calibration matrix (see $T^H_C$ in Figure \ref{fig:transform_diagram})
    \item Align medical probe to virtual drilling axis with the aid of AR assistance
    \item Confirm when satisfied that both the tool-tip position and tool orientation are aligned with displayed axis, pose of tool is then recorded for offline analysis
\end{itemize}

The virtual drilling axis was composed of a holographic cylinder and sphere to assist users with achieving the desired position and orientation for the planned tool trajectory. The aim of the task was to align the shaft of the medical probe with the holographic cylinder (locking rotation), whilst best aligning the tip of the probe with the centre of the holographic sphere (locking position).

During the alignment task, as both the probe and femur were continually tracked, it was possible to estimate a `live' rotational and translation error of the achieved tool-trajectory with respect to the plan, based on their computed virtual locations. Error magnitudes were used to modulate the colour of the virtual drilling axis from red to green to indicate when the achieved pose generated translation and rotational errors under a preset threshold ($<\,$5 mm, 5$^{\circ}$). This was intended as a coarse guidance mechanism, with users having the final say on when the tool was satisfactorily aligned with the target holographic trajectory. When the tool was both incorrectly aligned and positioned, both the cylinder and sphere appeared red. Once the tooltip positional error was under the threshold, the colour of the sphere began modulating from red to green. A similar process was carried out to modulate the colour of the cylinder from red to green when the angular error (computed using the dot product between the planned and estimated live trajectory) fell below the preset threshold. This modulation was linearly proportional to the magnitude of error in both instances, with Figure \ref{fig:trackingScenarios} showing different modulation scenarios.  
Once volunteers were satisfied with the alignment, they provided verbal confirmation, and the optical tracker recorded the achieved pose of the tool, consequently allowing for the calculation of a translation and rotation error when compared against the planned drilling axis. Each volunteer then repeated the alignment task for a total of 4 unique drilling trajectories.  

%%%%%%%%%%%%%%%%%% RESULTS
\section{Results}
\label{section:results}
\subsection{Relative Tracking Error}
\label{subsection-relative-track-results}
\begin{figure}%[H]
\vspace*{6px}
    \centering
    \includegraphics[width=.98\linewidth]{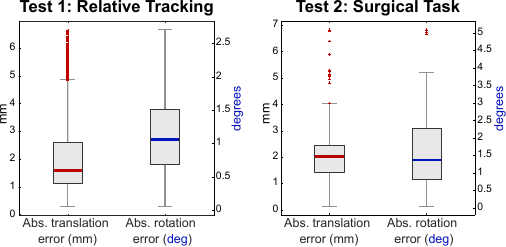}
    \caption{Box-whisker plots of absolute error magnitude for translation and rotation in the relative-tracking and surgical task tests. Whiskers extend to the data points not considered outliers (approximately $\pm\,$2.7$\sigma$).}
    \label{fig:boxplot_relativeError}
\end{figure}

As discussed in Section \ref{section:matMethods}, by comparing the pairs ($N$ = 7000) of measured and ground-truth relative-pose matrices, it was possible to calculate a 3D error for translation and rotation for relative tracking of IR-marker arrays. As the directionality of error was not of significance in this study, the mean-absolute-errors (MAEs) computed for translation and rotation were analysed and are reported in this study. The MAEs for translation and rotation were found to be 2.028 mm and 1.122$^{\circ}$ respectively. Aggregated statistics for the two error categories are reported in Table \ref{table:relativeError}-B, and are illustrated with a box and whisker plot in Figure \ref{fig:boxplot_relativeError}.

\begin{table}
  \caption{\textbf{(A)} Aggregated subject metadata \textbf{(B,C)} Aggregated Statistics for Relative Tracking Error and Surgical Task Tests}{\label{table:relativeError}}
  \centering
      \begin{tabularx}{0.9\linewidth}{Xll}
      \hline 
      \multicolumn{3}{c}{\textbf{(A) Subject Metadata}}	\\	\hline		
      N  & \multicolumn{2}{c}{8} \\ 
      Age (Mean, S.D) & 28.6 & 4.72 \\ 
      Sex (Male, Female) & 6 & 2 \\ 
      Prior experience in AR (Yes, No) & 3 & 5 \\ \hline
      \end{tabularx}
      \begin{tabularx}{0.9\linewidth}{Xccc}
      \hline 
      \multicolumn{4}{c}{\textbf{(B) Test 1 - Relative Tracking Errors}} \\ \hline 
      
      \hline
      & \textit{Mean} & \textit{SD} & \textit{Median}\\ 
      Translation (mm)  & 2.028  & 1.275 & 1.599 \\ 
      Rotation (deg) & 1.122 & 0.553 &  1.062 \\\hline

      \multicolumn{4}{c}{\textbf{(C) Test 2 - Surgical Task Errors}} \\ \hline 
      
      \hline
      & \textit{Mean} & \textit{SD} & \textit{Median}\\ 
      Translation (mm)  & 2.067  & 0.933 & 2.011 \\ 
      Rotation (deg) & 1.539 & 0.952 & 1.367\\\hline
      
      \hline
      \end{tabularx}
\end{table}

 %\vspace*{-1cm}
\subsection{Surgical Task Error}
\label{subsection:taskerror-results}
As each of the 8 volunteers carried out the tool alignment task described in Section \ref{section:matMethods}, the optical tracker took 64 readings per subject (4 unique trajectories $\times$ 16 readings per trajectory) of the achieved tool pose. A total of 512 (64 readings $\times$ 8 users) achieved tool poses were compared to their corresponding planned trajectories to compute a translational and rotational error. The MAEs for translation and rotation were 2.067 mm and 1.539$^{\circ}$ respectively. Aggregated statistical data for these results ($N$ = 512) are included in Table \ref{table:relativeError}-C, with box and whisker plots for the study results shown in Figure \ref{fig:boxplot_relativeError}.  

%%%%%%%%%%%%%%%%%%%%%%%%%% DISCUSSION
\section{Discussion}
In this study, we report the setup of an IR-based calibration system designed to co-register a surgical robot's optical tracker with a HoloLens 2 headset. Study tests investigated the relative tracking accuracy achieved by the reported system when tracking multiple IR-reflective marker arrays, as well as the accuracy achieved in a simulated wire-insertion task assisted by holographic guidance following a calibration between the AR headset and surgical robot. As previously highlighted, most existing studies to date achieve registration between surgical tracking systems and AR-HMDs through manual or image-marker based techniques. In comparison to manual techniques, our setup eliminates any time-consuming subjective calibration protocols, and in contrast to image-marker centric techniques, our system does not introduce new markers or tools to a surgical system's workflow, and solely uses IR-reflective marker arrays which are already part of the surgical scene in CAOS.

The results of the first test demonstrated the tool-tracking system discussed in this paper produced MAEs for relative tracking of 2.028 mm and 1.122$^{\circ}$ in a static setup. For context, optical tracking systems used in navigated surgery typically report sub-degree, sub-millimetre levels of accuracy \cite{Elfring2010AssessmentSystems}. Whilst these results do not achieve this benchmark, they are close and within the same order of magnitude.

The most comparable studies for contextualising the results of the relative-tracking test described in Section \ref{subsection:experiments} are prior works which investigated the HoloLens's ability to natively track surgical instruments equipped with IR-reflective markers. 
Gsaxner et al. {\cite{Gsaxner2021Inside-outReality}} achieved instrument-tracking using a bespoke setup where an IR-emitting rig was mounted to a HoloLens 2 to make IR-reflective markers in the scene fluoresce. Marker-array pose was obtained by processing filtered and paired grayscale images returned by the headset's front-facing visible light cameras. The authors reported static tracking RMS errors of 1.70$\,\pm\,$0.81 mm and 1.11$\,\pm\,$0.39$^{\circ}$ {\cite{Gsaxner2021Inside-outReality}}, figures comparable to our results. In comparison, we believe our setup could be more easily integrated with CAOS platforms, as the use of additional IR-emitters in the surgical scene alongside the HoloLens may cause undesired interference if any optical-tracking systems are in use. Additionally, our system represents a more plug-and-play solution for this scenario, with no additional modifications to the HMD, with the only requirement being a list of marker-array geometries tracked by the CAOS system. 
Kunz et al. {\cite{Kunz2020InfraredInterventions}} previously explored inside-out tracking of IR-reflective markers using the HoloLens 1 with two approaches: (1) grayscale stereo image analysis (similarly to Gsaxner et al. {\cite{Gsaxner2021Inside-outReality}}) (2) IR and ToF depth-image analysis, analogous in approach to the system reported in this paper. The authors investigated relative 3 DOF translation errors, reporting an error of 0.69 mm {\cite{Kunz2020InfraredInterventions}} for tracking individual IR reflective spheres, and did not provide errors for estimating 6 DOF pose of marker-arrays.
%, as was investigated during our study.
When compared to these recent studies, our system extends on the reported approaches, which enable inside-out tracking of surgical instruments, and uses this information to bring an AR-HMD into the navigation loop of commercial CAOS platforms utilising optical tracking systems. The challenges of occlusion and line-of-sight in traditional CAOS tracking setups using a single, static tracking device can be addressed, through the hybrid-tracking approach taken by our system, which co-registers an optical tracker with an HMD worn by a mobile user, and the calibration system can be used to support AR applications for CAOS which employ patient-registered, in-situ holographic guidance (see Figure {\ref{fig:ar_caos}). 
}

The second component of this study was a simulated wire-insertion task which built on the calibration setup. Users were provided overlaid holographic guidance (as illustrated in Figure \ref{fig:trackingScenarios}) designed to assist with correctly orienting and positioning a tool to best replicate a pre-computed tool trajectory. For this freehand task, the MAEs for translation and rotation achieved were 2.067 mm and 1.539$^{\circ}$ respectively. Whilst various HMD based studies in literature report task-based errors, results are often task and condition-specific, which may make direct comparisons of task-based errors unsuitable. For additional context, the reader is referred to a review by Andrews et al. \cite{Andrews2021RegistrationDevices}, which reports co-registered AR setups for surgical and medical applications in literature, and their associated task-based errors when using AR to guide or assist medical tasks; reported mean error values in these studies fall within a range of 1.07\,-\,5.76 mm, and 0.97\,-\,4.3$^{\circ}$. 

The limitations of this study can be considered in two groups: system-design and  the experimental conditions of the surgical task. As part of the system design, both registration and the static relative-tracking setup in the first test required a specific condition -- IR markers were only tracked when inside a 1 m range of the headset, due to a limitation set by the headset's API. Whilst this did not prevent the positioning of holographic objects, as once registration between the headset and optical tracker was achieved, holographic content could be positioned using optical tracker data, it meant continuous tracking of IR markers by the HMD was only possible in a smaller working volume when compared to modern optical trackers.

As the study's calibration system (used in both tasks) relies on the HMD's ability to locate itself in a virtual world-frame, the static design of the first test means further investigation is required to investigate how more dynamic relative-tracking setups are impacted by drift or accumulated errors generated by the HoloLens's inside-out tracking. Previous studies have investigated temporal drift associated with self-locating algorithms running on the HoloLens {\cite{Vassallo2017HologramHoloLens}}, {\cite{ Guinet2019ReliabilityConditions}}, however, in our application, any drift-errors associated with the virtual world frame can be easily addressed by the user looking at the calibration target or patient anatomy to re-calibrate the spatial transformation between the optical tracker and virtual world-frame in a matter of seconds.

During calibration, the latest captures of tool-pose calculated by the optical tracker and HMD at a particular instance were used to co-register the two systems. As calibration and the relative-tracking tests were carried out with stationary targets whilst trying to best minimise large, sudden motions of the head, the setup was less vulnerable to time-dependent errors when comparing the datastreams from the HMD and optical tracker. However, this remains a limitation, and future work will implement timestamp-based synchronisation to support more dynamic tracking scenarios. 
Additionally, the pipeline for tool-detection and tracking (see Figure {\ref{fig:img_proc}}) inherently used camera parameters (intrinsic and extrinsic matrices) of the HMD's depth-sensor, provided by the Research Mode API. Using manufacturer supplied parameters did not present significant errors for tool-tracking during use or when examining test results. However, whilst our system did not employ additional offline calibration steps to recalculate camera properties of the depth sensor, the accuracy of future iterations of the system could benefit from additional camera calibration processes to verify properties of the depth sensor, at the cost of additional setup time and making the setup more device-specific.

With regards to experimental condition limitations, the second task -- a simulated wire insertion into the distal femur, is not a commonly practised orthopaedic procedure. This task was also carried out by users in a controlled laboratory setting on a clean and isolated plastic femur model, without the presence of challenging factors associated with typical surgical scenes such as surgical lighting, soft tissue, blood or medical drapes \cite{Liu2018AugmentedStudy}. 
\section{Conclusion}
We propose a calibration system for rigidly registering a HoloLens 2 headset with an optical tracker used by a surgical robot. By fusing IR response and near-depth data provided by sensors on board the HoloLens 2, a customised app deployed to the headset was able to optically track IR-reflective marker arrays simultaneously tracked by the surgical robot. The system reported tracking MAEs of 2.028 mm and 1.122$^{\circ}$ when using the HoloLens 2 to track relative pose between two marker-arrays in a static setup. In a second experiment, in-situ holographic guidance was generated by the calibration system to assist ten volunteers with a simulated distal-femur wire-insertion task, producing MAEs of 2.067 mm and 1.539$^{\circ}$ when compared to a target trajectory. 
The results of these tests indicate a level of accuracy, which whilst not yet clinically acceptable ($<\,$1 mm, $<\,$1$^{\circ}$), is close to this threshold. 
Overall, the system presented in this paper provided a fast, user-agnostic method for calibrating an AR headset with a surgical robot, and will be utilised in future work that aims to develop a new mixed-reality workflow for robot-assisted surgical procedures.

\section*{Acknowledgments}
The authors would like to thank Jason Cipriani, Dylan Hower, and Dr Fabio Tatti for their roles in the development of a research edition of the \navio\ robot's software. The design of the C++ DLL constructed for this study was based on an open-source project originally made available at \cite{Gu2021HoloLens2-ResearchMode-UnityCode}. 
\bibliographystyle{IEEEtran}
\bibliography{references}
\end{document}